\pdfoutput=1
\documentclass[11pt]{article}

\usepackage[]{autoeval2021}

\usepackage{times}
\usepackage{latexsym}
\usepackage{enumitem}
\usepackage[T1]{fontenc}
\usepackage[utf8]{inputenc}

\usepackage{microtype}

\usepackage{subcaption}
\usepackage{booktabs}
\usepackage{tabularx}
\usepackage{arydshln}
\usepackage{multirow}
\usepackage{graphicx}
\usepackage{array, makecell}
\usepackage{amssymb}
\usepackage{xcolor} 
\usepackage{soul}
\usepackage{titlesec}

\titlespacing*{\section}
{0pt}{2.5ex plus 1ex minus .2ex}{1.25ex plus .2ex}
\titlespacing*{\subsection}
{0pt}{1.5ex plus 1ex minus .2ex}{1.0ex plus .2ex}

\title{What is wrong with you?: Leveraging User Sentiment for Automatic Dialog Evaluation}

\author{
Sarik Ghazarian$^{1}$\thanks{~ Work done while Sarik Ghazarian was an intern at Amazon Alexa AI} \quad Behnam Hedayatnia$^{2}$ \quad Alexandros Papangelis$^{2}$ \\ \textbf{Yang Liu$^{2}$ \quad Dilek Hakkani-Tur$^{2}$}\\
$^1$ University of Southern California / Information Sciences Institute\\
$^2$ Amazon Alexa AI\\
\small{\texttt{sarik@isi.edu}} \\
\small{\texttt{\{behnam,papangea,yangliud,hakkanit\}@amazon.com}}
}

\begin{document}
\maketitle
\begin{abstract}

Accurate automatic evaluation metrics for open-domain dialogs are in high demand. Existing model-based metrics for system response evaluation are trained on human annotated data, which is cumbersome to collect. In this work, we propose to use information that can be automatically extracted from the next user utterance, such as its sentiment or whether the user explicitly ends the conversation, as a proxy to measure the quality of the previous system response. This allows us to train on a massive set of dialogs with weak supervision, without requiring manual system turn quality annotations. Experiments show that our model is comparable to models trained on human annotated data. Furthermore, our model generalizes across both spoken and written open-domain dialog corpora collected from real and paid users. 
%Finally, we show that user sentiment can be used for estimating the overall dialog quality.

% Existing metrics mainly focus on turn-level evaluations and use simple aggregation techniques to compute a dialog-level score. 
%Additionally, we show that the position of the turns can be a useful feature for estimating the overall dialog quality.

%and show our model 
% outperforms previous automatic evaluation models in both domains.  

%Specifically we propose three methods: one that predicts the next sentiment directly, and two others that predict the next user utterance using an utterance or a feedback generator model and then classify its sentiment.

\end{abstract}

\section{Introduction}
\label{sec:introduction}
Relying on human evaluation to determine the quality of open-domain dialog systems is not an efficient approach in terms of time and cost.  
%rely on human evaluation to determine the quality of the system which is time-consuming and expensive. 
%Additionally most human evaluation is done using paid participants which can lead to inconsistent results as these participants may not actually care about the annotation task. 
%Recently open-domain dialog systems have relied on end-to-end generation models~\cite{adiwardana2020towards, roller2020recipes, dinan2018wizard, gopalakrishnan2019topical, bao2019plato}. 
Automatic evaluation can be a good replacement for human annotations and can increase the pace of open-domain dialog system development. However, standard word-overlap metrics (BLEU, ROUGE, Perplexity) do not correlate well with human judgements of open-domain dialog systems~\cite{deriu2020survey, liu2016not} because of the diverse set of outputs that can be relevant given a dialog context. 

%Automatic evaluation for open-domain systems can be broken down into reference-based and reference-free metrics. Reference-based metrics~\cite{lowe2017towards, tao2018ruber, ghazarian2019better, lan2020pone, zhang2021deep, xiang2021assessing} require at least one ground truth response to evaluate a response against; however, it can be expensive to collect multiple ground truth responses as there are many valid responses at a given turn within a dialog. Reference-free metrics~\cite{pang2020towards, sinha2020learning, huang2020grade, ghazarian2020predictive, mehri2020usr, zhang2021dynaeval, li2021conversations, mehri2020unsupervised} do not require a ground truth response for evaluation.

% BEHNAM
%TODO: Make sure all automatic eval papers are cited

A solution for better automatic evaluation methods is to train reference-free evaluators that learn how to assess the generated responses given dialog contexts from different aspects 
%Most automatic evaluation metrics focus on fine-grained assessment of response quality looking at measures
such as relevancy~\cite{tao2018ruber, ghazarian2019better, lan2020pone}, engagement~\cite{ghazarian2020predictive}, fluency~\cite{zhang2021deep, pang2020towards}, contradiction~\cite{pang2020towards, nie2021like} amongst others. It is also important to get some holistic evaluation at the dialog level in order to assess the dialogs as a whole~\cite{zhang2021dynaeval, li2021conversations, mehri2020unsupervised, finch2021went}. %We leverage features from user utterances and show through simple aggregation that we can automatically estimate dialog level scores. 

%However, not every turn within a dialog contributes equally to its overall rating. Similar to~\cite{li2021conversations, finch2021went} for dialog level rating estimation we do not give the same importance to each turn. We provide an analysis from human annotations confirming this hypothesis.

Recently,~\citet{mehri2020unsupervised, eskenazi2019beyond} have shown the effectiveness of looking into the next user utterance as a proxy to measure the quality of the chatbot's generated responses. 
\citet{see-manning:2021:sigdial} have shown that predicting next user satisfaction helps to select more relevant system utterances. Inspired by works in this area, we propose to automatically extract features from the next user utterance, such as sentiment, to use as a proxy to evaluate system responses. The advantage of our method is that we do not need to train on data with human annotations for turn level quality, and instead can rely on available large datasets with automatically extracted annotations.
%of the generated responses.
%Indeed, they use the likelihood of a set of predefined positive and negative potential user responses based on DialoGPT~\cite{zhang2019dialogpt}.

Most existing automatic evaluators focus purely on open-domain text-based dialog systems. In addition to textual interactions, we perform experiments on  voice-based interactions that were collected via paid and real users. Furthermore, we compute correlations with a real user's own (referred to as first party, 1P) rating when available, in addition to annotations by third party (3P) annotators.
Our contributions include:
% \vspace*{-1.5ex}
\begin{enumerate}
    \itemsep -0.75ex
    %\item a next user response and feedback generator and next user sentiment estimator to evaluate open-domain dialogue systems at both dialog and turn levels,
    \item training an automatic evaluator on the sentiment of the next user utterance in a weakly supervised fashion to evaluate system responses,
  %  \item estimating dialog level scores by aggregating turn level sentiment,
    %\item leveraging turn positions to better estimate dialog level scores,
    \item outperforming existing automatic evaluation metrics on both text and voice-based open-domain dialog datasets,
    \item a turn-level annotated open-domain text-based dialog dataset that we will release.\footnote{We cannot release our voice-based interactions due to privacy concerns that will be discussed in the paper.}
\end{enumerate}

\begin{figure}[t]
 	\centering
 		\includegraphics[width=0.5\textwidth]{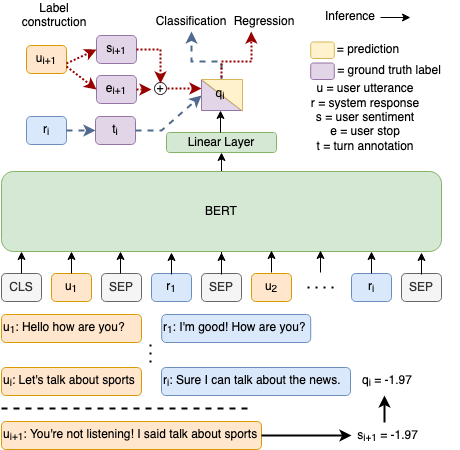}
 	\caption{Training/Inference for turn quality estimation.  The dotted arrows show how $q_i$, which represents the system turn quality for system response $r_i$, is constructed for training. For our regression model indicated by the red arrow, $s_{i+1}$ (user sentiment) and $e_{i+1}$ (user stop) are summed together to create $q_i$. For our classification model indicated by the blue arrow, $q_i$ is equal to $t_i$. In the example dialog, the user expresses negative sentiment in $u_{i+1}$. The sentiment score -1.97 is used as the reference label $q_i$,  indicating the quality of response $r_i$\vspace*{-1.5ex}.}
	\label{fig:nus_model}
\end{figure}

%\vspace{-3em}
\section{Methods for Automatic Evaluation}
%\vspace{-1em}
%\subsection{Turn level quality}
%\subsection{Turn Quality Estimation}
%Based on the feature analysis done in Section~\ref{sec:feature} we explore three setups: Next user quality estimation, utterance generation, and feedback generation.
For turn quality estimation, the task is defined as follows:  given a dialog context and a system response in the last turn, $D$ = [$u_1$, $r_1$ \dots $u_i$, $r_i$] (where $u_i$ and $r_i$ are the user utterance and system response respectively for the $i^{th}$ turn in a dialog), determine if $r_i$ is an appropriate response. $q_i$ indicates the quality of response $r_i$ and will be used as our reference label when training the model.
Figure~\ref{fig:nus_model} shows our model architecture. We train a BERT-base~\cite{devlin2018bert} model that encodes the dialog context and the latest system response.  We use the pooled representation output by the BERT model and pass it through a linear layer to determine the quality of the response. Depending on the reference label used to train this model, we adopt a classification or regression setup, described below. 

\begin{itemize}
\itemsep -1ex
    \item {\bf Classification model trained using turn level annotations.}
When annotations for system responses are available in our training data (a binary label $t_i$ as shown in Figure~\ref{fig:nus_model} for response $r_i$, indicating if the system response is appropriate), we train a classification model using such reference labels.   
  
  \item
{\bf Regression model trained using next user sentiment.} 
Obtaining turn level annotations for dialogs is costly. In this work, we explore using weak supervision to approximate response quality.
\citet{eskenazi2019beyond} stated that given a system response, the follow up user's utterance should be used to evaluate the quality of the system response as it increased agreement amongst human annotators. Motivated by this, we propose to use the sentiment of the next user utterance as a proxy to estimate the quality of the previous system response. In Figure~\ref{fig:nus_model}, $s_{i+1}$ is the sentiment score for the next user utterance $u_{i+1}$. 
%We run two experiments where the sentiment score is used as the target label and where the sentiment score is combined with~\textbf{Next user stop signal} as a target label which is described below. 
Note that this information is automatically generated from the user utterance, and thus allows us to leverage data without a turn level annotation. 
Since such sentiment scores are often continuous, we use a regression model for these target labels.

\item
{\bf Next user stop signal.}
We also examine if the next user utterance stops a dialog ($e_{i+1}$ in Figure~\ref{fig:nus_model}). $e_{i+1}$ is 0 if the user stops the dialog and 1 if they continue the dialog. We use this as an additional signal by summing it with the sentiment information above as target labels for model training.

\end{itemize}

\vspace*{-1.5ex}
For dialog level evaluation, 
we follow previous work and use mean aggregation techniques to estimate dialog level ratings from turn level scores~\cite{lowe2017towards, ghazarian2019better, ghazarian2020predictive, lan2020pone, yeh2021comprehensive}. 
%Because turn level scores need to be manually annotated we automatically tag user utterances with sentiment scores. 
In our experiments, we show how aggregated turn level quality  and user sentiment scores correlate with dialog level ratings. 

% BEHNAM
%\subsection{RUBER}

%\subsection{BERT-RUBER}

%\subsection{PONE}

%\subsection{PredictiveEng}

%\subsection{FED}

\begin{table*}[h]
\centering
\small
\begin{tabular}{c c c c c c}
 \hline
Dialog Split 
& \makecell{Number of Interactions \\ (Train/Dev/Test)} 
& \makecell{Avg. Number of Turns \\ (Train/Dev/Test)}
& 3P turn quality
& 3P rating
& 1P rating \\
%& 1P feedback\\
%\hdashline[.4pt/1pt]
~\textit{PUI}  & -/-/87 & - / - / 14.5 & \checkmark & \checkmark & \\
\hdashline[.4pt/1pt]
%~\textit{RUI-small} & 2717 / 735 / - & 10.1 / 10.7 / - &  & & \checkmark & \checkmark \\
\hdashline[.4pt/1pt]
~\textit{RUI-1P} & 6215 / 690 / - & 10.3 / 10.8 / - &  & & \checkmark \\
\hdashline[.4pt/1pt]
%~\textit{RUI} & 79338 / 19834 / - & 9.8 / 9.8 / - & &  & \checkmark \\
\hdashline[.4pt/1pt]
~\textit{RUI-3P} & 500 / 55 / 132 & 11.1 / 10.7 / 14.3 & \checkmark & \checkmark & \checkmark \\
\hdashline[.4pt/1pt]
~\textit{ConTurE} & - / - / 119 & - / - / 8.95 & \checkmark & \checkmark\\
\hline
\end{tabular}
%\caption{Dataset Statistics for Spoken dialog dataset\vspace{-3ex}. RUI (Real User Interactions)}
\caption{Dataset Statistics for Spoken and Written dialog datasets\vspace{-3ex}. RUI (Real User Interactions)}
\label{prize}
\end{table*}

\vspace{-0.9em}
\section{Dialog Datasets}
As described earlier, most previous work in automatic evaluation focuses on text-based open-domain dialog systems~\cite{yeh2021comprehensive, lan2020pone, sinha2020learning, huang2020grade, ghazarian2020predictive}. Additionally most dialog datasets are collected via crowdworkers. While we also evaluate on written (text-based) dialogs, the primary dataset in our work consists of spoken (voice-based) interactions between a dialog system and a real user.

\begin{figure*}
 	\centering
 		\includegraphics[width=1.0\textwidth]{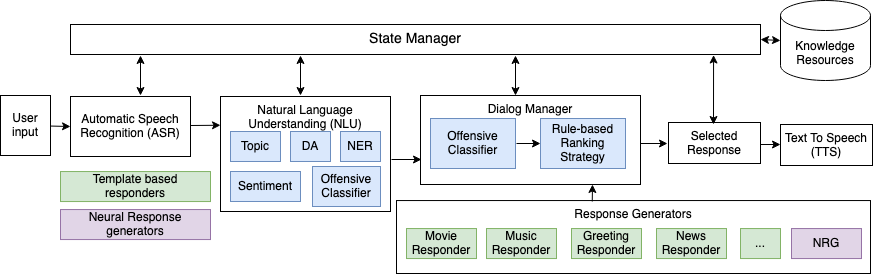}
 	\caption{Architecture of our open-domain dialog system. NER = Named Entity Recognition, DA = Dialog Act}
	\label{fig:dialog_system}
\end{figure*}

\subsection{Open Domain Dialog System}
\label{sec:dialog_system}

We first describe the open-domain dialog system used for our spoken dialog data collection. The architecture of our dialog system is shown in Figure~\ref{fig:dialog_system}.
Every user utterance in the dialog is sent into an ASR system whose output goes through a series of NLU modules that classifies topics, dialog acts, sentiment, extracts entities, and detects if the user utterance is offensive. Our system then calls multiple response generators (called responders) for the given dialog context and logs all the generated response candidates within the State Manager. The final response is selected based on a rule-based ranking strategy, and then sent to the TTS module whose output is presented to the user.

For popular topics in open domain dialogs, such as movies, music, recent news, we develop template-based responders (highlighted in green in Figure~\ref{fig:dialog_system}) for the given dialog state. 
An example state and response for the movie domain is: when the user turn mentions a movie name (based on the NER result), we respond with information about the actor, the rating, or the plot of this certain movie. In addition to topic-specific template-based responders, our system includes other template-based responders for some special dialog contexts, such as greetings, topic switches, etc. 

For every user turn, we also apply a neural network-based response generation (NRG) model to produce a response, highlighted in purple in Figure~\ref{fig:dialog_system}. 
Our~\textit{NRG Responder} is a GPT2-XL~\cite{radford2019language} based model trained on real user-system interactions described in Section~\ref{sec:data}. 

The rule-based response ranker uses predefined logic and selects a template-based responder when it is available and the user topic matches that responder, otherwise it uses the NRG response as a fall back. In our system since we have just a few template-based responders, the system uses NRG responses most of the time. 

\subsection{Spoken Dialogs}
\label{sec:data}

We deploy the dialog system described above  within the Alexa Prize Socialbot framework~\cite{ram2018conversational} to interact with real users.
A user initiates an interaction with our dialog system and consents to have their data collected. A turn within an interaction is specified as a user utterance-system response pair. These interactions end when the user requests to stop the conversation. At the end of each interaction, users are given the opportunity to leave a rating in the range of 1 to 5. We define these ratings as~\textit{1P rating} as they come from the same users who interacted with the conversational agent. We denote this dataset as~\textit{Real User Interactions (RUI)}\footnote{All interactions are in English.}. Our data consists of approximately 100k interactions and 5 million turns. This dataset is used to train our~\textit{NRG Responder} mentioned in the previous section. We discuss its training details  in the Appendix. 

Not every user leaves a rating; therefore, we take a sample of interactions from~\textit{RUI} that contain user ratings and denote this dataset as~\textit{RUI-1P}.

%This collection contains 6905 interactions with an average length of 10.8 turns.

In addition to real user interactions, we form a dataset of interactions from paid users who were instructed to speak to the same dialog system. We denote these interactions as~\textit{paid user interactions PUI}$^2$. The difference between paid and real users is that the former are internal workers who are recruited to rigorously test and probe the dialog system and as a result are much more proactive in the dialogs as opposed to real users who are known to be less proactive in these social conversations~\cite{juraska2021athena, finch2020emora}. These internal workers are considered paid as their primary job consists of assisting with data collection. Real users, however, are consenting to a dialog with our dialog system but are not being paid.

%This has 87 interactions with 14.5 turns on average.

To obtain turn quality labels, we annotate a subset of ~\textit{RUI-1P} at the turn level. Given a complete interaction, an experienced annotator was asked to annotate each system response either as 1 or 0, where 1 indicates the response is appropriate and vice versa for 0. Additionally, we ask annotators to leave a dialog level rating in the range of 1 to 5. We define this turn and dialog level annotations as~\textit{3P turn quality} and~\textit{3P ratings} respectively, since they came from annotators who rated other users' interactions. We denote this annotated data as~\textit{RUI-3P}. An example of a turn level annotation is shown in the Appendix. We also perform the same annotation on the~\textit{PUI} data. Table~\ref{prize} shows the statistics for each of these collections and available annotations for each dataset.\footnote{We cannot release this data publicly as it is real user data.}

%(687 interactions, average 11.7 turns)

%In the case of an issue, annotators were asked to select an error category for the issue in the system turn.

To obtain sentiment labels, we leverage the BiLSTM sentiment model from~\cite{kim2020speech}, which was trained on spoken dialog data and automatically tag user utterances with sentiment. The model takes in both audio and textual features and outputs a real-valued valence score on a scale from -3 to 3, which measures the degree of the utterance's positivity/negativity.

\subsection{Written Dialogs}
% https://drive.google.com/file/d/1oVDMg-6HffMpkSXWylafTvVI1pDuMuqw/view

%Currently the only human-system dialog dataset that has both turn and dialog level annotations is the ConvAI dataset~\footnote{http://convai.io/2017/data/}. However~\citet{logacheva2018dataset} shows that 58.6\% of system utterances were labeled inappropriate which leads to an unbalanced dataset. 
%We created a dataset that has an almost equal distribution of appropriate and inappropriate turns to evaluate our model. 
We sample a set of dialogs released from the Interactive Evaluation of Dialog track~\cite{gunasekara2020overview} to be annotated for turn quality. These dialogs were collected from invited participants conversing with knowledge-grounded response generation models through textual exchanges, and have been publicly released\footnote{https://github.com/exe1023/DialEvalMetrics}. The original authors of this dataset asked Amazon Mechanical Turk (AMT) workers to rate 2200 interactions on multiple dialog level dimensions, such as coherent, informative, overall. The full list of dialog level annotation dimensions is included in the Appendix. However, these dialogs do not have turn level annotations. In order to evaluate our models at the turn level, we sample 119 dialogs with an average length of 8 turns. For each turn, we ask three AMT workers to rate whether they dislike, somewhat like or like the Chatbot's response with a score of either 0, 1, or 2 respectively. To help workers judge response quality, we ask them to look at how relevant and interesting a response is. We use majority voting to determine the final score. In the case of ties we use a score from an internal author. The Krippendorff's alpha score is 0.31 representing fair agreement between annotators. We denote these assessments as~\textit{3P turn quality} since the AMT workers are rating other workers' dialogs. We denote this dataset as~\textit{Conversational Turns Evaluation (ConTurE)} and publicly release it.\footnote{We release the ConTurE dataset at \url{https://github.com/alexa/conture}}

\begin{table*}
\centering
\small
\begin{tabular}{c c c c c c c c}
 \hline
Training Set
& Model (Ref label)
& \multicolumn{2}{c}{RUI-3P (test set)}
& \multicolumn{2}{c}{PUI}
& \multicolumn{2}{c}{ConTurE}\\
\toprule
&
& Pearson  & Spearman
& Pearson  & Spearman 
& Pearson  & Spearman \\
\hline
%- & RUBER & -0.08 & -0.07 & -0.01 & -0.03\\
- & RUBER & -0.08 & -0.07 & -0.1 & -0.1 & -0.01 & -0.03\\
%- & RUBER & -0.08 & -0.07 & -0.01 & -0.03\\
\hdashline[.4pt/1pt]
%- & BERT-RUBER  & 0.01 & 0.02 & -0.007 & 0.004\\
- & BERT-RUBER  & 0.01 & 0.02 & -0.02 & -0.02 & -0.007 & 0.004\\
%- & BERT-RUBER  & 0.01 & 0.02 & -0.007 & 0.004\\
\hdashline[.4pt/1pt]
%- & PONE & 0.01 & 0.004 & 0 & 0.01 \\
- & PONE & 0.01 & 0.004 & -0.02 & -0.03 & 0 & 0.01 \\
%- & PONE & 0.01 & 0.004 & 0 & 0.01 \\
\hdashline[.4pt/1pt]
%- & PredictiveEng & -0.11 & -0.11 & -0.11 & -0.09\\
- & PredictiveEng & -0.11 & -0.11 & -0.06 & -0.05 & -0.11 & -0.09\\
%- & PredictiveEng & -0.11 & -0.11 & -0.11 & -0.09\\
\hdashline[.4pt/1pt]
%- & FED & -0.006 & -0.02 & 0.11 & 0.10\\
- & FED & -0.006 & -0.02 & -0.03 & -0.04 & 0.11 & 0.10\\
%- & FED & -0.006 & -0.02 & 0.11 & 0.10\\
%\hdashline[.4pt/1pt]
\hline
\multicolumn{8}{c}{Our method} \\ \hline
%~\textit{RUI-issue} & Classification (3P) &~\textbf{0.29} &~\textbf{0.28} & -0.01 & 0.11\\
RUI-3P & Classification (3P) &~\textbf{0.29} &~\textbf{0.28} & 0.23 & 0.24 & -0.01 & 0.11\\
%~\textit{RUI-issue} & TQP (3P) &~\textbf{0.29} &~\textbf{0.28} & -0.01 & 0.11\\
\hdashline[.4pt/1pt]
%~\textit{RUI} & Reg (Sentiment) & 0.15 & 0.12 &~\textbf{0.34} &~\textbf{0.34}\\
RUI-1P & Reg (Sentiment) & 0.15 & 0.12  & 0.19 & 0.16 &~\textbf{0.34} &~\textbf{0.34}\\
%~\textit{RUI} & NUS (Sentiment) & 0.15 & 0.12 &~\textbf{0.34} &~\textbf{0.34}\\
%\hdashline[.4pt/1pt]
%~\textit{RUI-issue} & TQP (Sentiment + 3P) & 0.24  & 0.22 & 0.27 & 0.27 & 0.29 & 0.29\\
%~\textit{RUI-issue} & TQP (Sentiment + 3P) & 0.24  & 0.22 & 0.29 & 0.29\\
%\hdashline[.4pt/1pt]
%~\textit{RUI-issue} & TQP (3P + User Stop) & 0.27 & 0.26 & 0.27 & 0.28 & 0.34 & 0.36\\
%~\textit{RUI-issue} & TQP (3P + User Stop) & 0.27 & 0.26 & 0.34 & 0.36\\
%\hdashline[.4pt/1pt]
%~\textit{RUI-issue} & TQP (Sentiment + 3P + User stop) & 0.25 & 0.22 & 0.26 & 0.25 &~\textbf{0.4} &~\textbf{0.4} \\
%~\textit{RUI-issue} & TQP (Sentiment + 3P + User stop) & 0.25 & 0.22 &~\textbf{0.4} &~\textbf{0.4} \\
\hdashline[.4pt/1pt]
%~\textit{RUI-small} & NUS (Sentiment + User Stop) & 0.2 & 0.22 & 0.31 & 0.32\\
\hdashline[.4pt/1pt]
%~\textit{RUI} & Reg (Sentiment + User Stop) & 0.22 & 0.23 & 0.3 & 0.33\\
RUI-1P & Reg (Sentiment + User Stop) & 0.22 & 0.23 &~\textbf{0.35} &~\textbf{0.3} & 0.3 & 0.33\\
%~\textit{RUI} & NUS (Sentiment + User Stop) & 0.22 & 0.23 & 0.3 & 0.33\\
\hdashline[.4pt/1pt]
%~\textit{RUI (40\%)} & Reg (Sentiment + User Stop) & 0.2 & 0.22 & 0.31 & 0.32\\
RUI-1P (40\%) & Reg (Sentiment + User Stop) & 0.2 & 0.22 & 0.29 & 0.24 & 0.31 & 0.32\\

\hline
\end{tabular}
\caption{Correlation between both baseline and our model outputs against~\textit{3P turn quality} for spoken and written datasets. For our method, reference labels used for Classification or Reg (Regression) models are presented.\vspace{-3ex}}
\label{baseline_models}
\end{table*}

\section{Results and Discussions}
%\subsection{Baseline models}
We compare our method with a suite of open source models from~\cite{yeh2021comprehensive}$^4$ %\footnote{https://github.com/exe1023/DialEvalMetrics}
including RUBER, BERT-RUBER, PONE, PredictiveEngagement and FED~\cite{tao2018ruber, ghazarian2019better, lan2020pone, ghazarian2020predictive, mehri2020unsupervised}.

Table~\ref{baseline_models} shows the automatic turn level quality estimation, measured using both Pearson and Spearman correlation against turn level annotations on three datasets for different methods. 
On the spoken dialog testsets(\textit{RUI-3P} and~\textit{PUI}) the baseline models perform poorly. In contrast, our Classification(3P) model trained using~\textit{3P turn quality} achieves the highest correlation (0.29/0.28) on~\textit{RUI-3P}. This can be partly explained by the matched training and testing setup.
%however, this model was trained on turn level annotated data which can be costly to collect. 
%As a result we train on data that contains semi-supervised labels. 
We observe promising results for the Reg (Sentiment + User Stop) model which was trained with next user sentiment information combined with stop signal which achieves the highest correlation on the~\textit{PUI} test set and a correlation of (0.22/0.23) on~\textit{RUI-3P}. 
This demonstrates the effectiveness of weak supervision. We compare different training sizes~\textit{RUI-1P} (40\%) versus~\textit{RUI-1P} and show the expected benefit of more data for model training.    
% model trained on~\textit{RUI-small} can achieve positive correlation (0.2/0.22). Additionally if we increase the number of training dialogs and use~\textit{RUI} the correlation improves (0.22/0.23). These same results when testing on~\textit{PUI} where training NUS (Sentiment + User Stop) on more dialogs results in higher correlation (0.35/0.3). 
We also see that our models outperform the baseline models on the~\textit{ConTurE} testset. It is important to note that all the baseline models have been designed and evaluated using written dialogs, and though our models were fine-tuned only on spoken dialog, they are able to generalize to a different modality. FED has been shown to be a good dialog-level evaluator~\cite{yeh2021comprehensive}. However we see in Table~\ref{baseline_models} that FED achieves low performance for turn-level evaluation. This matches the conclusion in~\cite{mehri2020unsupervised} who point out that FED captures the dialog-level qualities from its training data Reddit better than turn-level qualities.

Table~\ref{conv_sentiment_correlation} shows the correlation results of the aggregated turn level scores with ~\textit{3P ratings} and~\textit{1P ratings} on the spoken dataset. 
From the first row, we can see that there is a moderate positive correlation between the aggregated mean of~\textit{3P turn quality} and~\textit{3P ratings} (0.50 / 0.46), but see a very low positive correlation with~\textit{1P ratings} (0.16 / 0.12). This may be due to the fact that Likert scale ratings can have lower inter-annotator agreement~\cite{belz2010comparing}. Additionally, the 3P annotators had access to the whole interaction and could re-read the context. 
This is in contrast to 1P users who may forget what happened earlier in the interaction as it is spoken.
Another reason is that 3P annotators only read the transcript of the dialog for turn or dialog evaluation, and may miss the tones in utterances that may affect 1P user ratings.   
When using the user sentiment scores, we can see through mean aggregation  it has positive correlation with both~\textit{3P ratings} (0.48/0.46) and ~\textit{1P ratings} (0.38/0.37). 
The higher correlation of user sentiment (as opposed to~\textit{3P turn quality}) with~\textit{1P rating} is partly because of the different signals used in 3P annotation as discussed above.
These results suggest sentiment can be used to estimate dialog level ratings, as done in previous work such as ~\cite{kim2020speech}. 

%\vspace*{-5ex}
Overall, we see that the next user utterance sentiment serves as a reasonable proxy to the quality of the previous system response, hence when there is not much data with turn level quality annotation, we can train models using weak supervision coming from the next user utterance.
In this study, we use the sentiment scores obtained from user utterances in speech based dialogs, therefore, acoustic features were used to obtain such sentiment information. 
Since speech based sentiment or emotion recognition has been widely studied, it does not require much additional annotation to train the sentiment model for user utterances, and thus we can rely on existing models. 
We also explored using sentiment just based on text, but observed some issues in our preliminary study. For example, when users reply with a `no' to a question, it is classified as negative, however, this may not indicate a problem with the previous system response. We plan to further investigate this in our future work, which will allow us to better utilize more available text based dialog data. 
Example outputs from both FED and our model are shown in the Appendix.

\vspace*{-2ex}

\begin{table}[h]
\centering
\small
\begin{tabular}{l c c c c}
    \toprule
     & \multicolumn{2}{c}{3P Ratings} 
     & \multicolumn{2}{c}{1P Ratings} \\ \toprule
    & P & S
    & P & S
    \\ \hline
    \multirow{1}{*}{3P turn quality} 
     &~\textbf{0.50} &~\textbf{0.46} & 0.16 & 0.12 \\
   \hdashline[.4pt/1pt]
    %\multirow{1}{*}{3P turn annotations (mean agg.)}
     %& - & 0.6/0.2/0.2 & 0.40 & 0.40 & 0.10 & 0.07 \\
    %\hdashline[.4pt/1pt]
    %\multirow{1}{*}{3P turn annotations (mean agg.)} 
     %& - & 0.2/0.2/0.6 & 0.50 & 0.43 & 0.20 & 0.16 \\
    %\hdashline[.4pt/1pt]
    %\multirow{1}{*}{3P turn annotations (mean agg.)} 
     %& - & 0.1/0.4/0.5 &~\textbf{0.51} & 0.45 &~\textbf{0.19} &~\textbf{0.15} \\
    %\hdashline[.4pt/1pt]
    %\multirow{1}{*}{3P turn annotations (mean agg.)} 
    %& - & 0.1/0.5/0.4 & 0.49 & 0.40 & 0.18 & 0.13 \\
    %\hdashline[.4pt/1pt]
    \multirow{1}{*}{User sentiment}  %& Satisfaction & 0.45 & 0.42 &~\textbf{0.34} &~\textbf{0.33}\\
     %& Satisfaction & 0.6/0.2/0.2 & 0.35 & 0.33 & 0.31 & 0.29 \\
     %& Satisfaction & 0.1/0.4/0.5 & 0.45 &~\textbf{0.43} & 0.31 & 0.30 \\
      %\hdashline[.4pt/1pt]
      & 0.48 & 0.46 &~\textbf{0.38} &~\textbf{0.37} \\
     %& Valence & 0.6/0.2/0.2 & 0.37 & 0.32 & 0.34 &  0.31 \\
      %& Valence & 0.1/0.4/0.5 &~\textbf{0.49} &~\textbf{0.48} & 0.35 & 0.36 \\
      %\hdashline[.4pt/1pt]
     %& Activation & 0.12 & 0.11	& 0.10 & 0.08 \\
     %& Activation & 0.6/0.2/0.2 &~\textbf{0.14} &~\textbf{0.12} & 0.09 & 0.06 \\
     %& Activation & 0.1/0.4/0.5 & 0.08 & 0.08 & 0.10 & 0.08 \\
    %\hdashline[.4pt/1pt]
   %\multirow{3}{*}{1P feedback sentiment} 
    % & Satisfaction & equal & 0.27 & 0.30 & 0.43 & 0.48 \\
    % & Valence & equal & 0.29 & 0.31 & 0.46 & 0.50 \\
     %& Activation & equal & 0.13 & 0.13	& 0.15 & 0.15 \\
    %\hdashline[.4pt/1pt]
    %RUBER & -0.02 & -0.02 & -0.07 & -0.07 \\
    %BERT-RUBER & 0.14 & 0.09 & -0.04 & -0.05 \\
    %PONE & 0.15 & 0.11 & -0.02 & -0.03 \\
    %PredictiveEng & 0.01 & 0.06 &~\textbf{0.25} &~\textbf{0.24} \\
    %FED & 0.22 & 0.20 & 0.10 & 0.10 \\
    %Classification (3P) & -0.04 & -0.05 & 0.12 & 0.15 \\
    %Regression (S) &~\textbf{0.27} & 0.30 & 0.13 & 0.16 \\
    %Regression (S + US) &~\textbf{0.27} &~\textbf{0.35} & 0.16 & 0.16\\
 \hline
\end{tabular}
\caption{\label{conv_sentiment_correlation} Correlation between turn level information (3P turn quality and user turn sentiment) and dialog level rating on RUI-3P. P=Pearson, S=Spearman.\vspace{-1.5ex}}
%equal=0.33/0.33/0.33} 
%The turn level scores are aggregated by mean}
\vspace{-1em}
\end{table}

\section{Conclusion}
In this work, we show that instead of training on manually annotated data we can train on sentiment from the next user utterance in a weakly supervised manner to evaluate system responses. We show that our model has better cross domain generalization and performs well on a written dialog dataset. 
In our future work we will investigate other methods beyond simple aggregation for dialog level estimation and using more text based dialog data. 
%Additionally we show that user sentiment can be used to estimate the overall dialog score through simple mean aggregation.

%Additionally we show the impact of turn positions in overall dialog evaluation and conclude that turns in the latter part of the dialog have more effect on the overall dialog score. 

%weshowed the impact of turn positions in overall dialog evaluation.
%showing that middle and ending turns in a dialogue are more important than beginning turns.
%We also explored the idea of looking into future turns and their sentiments to evaluate the quality of the generated responses. According to our findings, predicting users' next utterance sentiment scores have a positive impact on the accuracy of the evaluation metrics. 
%We also showed that even though predicting the sentiment of the generated feedback for the dialogue doesn't result the best correlation, its performance is very close to FED baseline with the main advantage of its ability to provide more detailed opinions about the quality of the dialogues that can be used for training better chatbots.

\section{Ethics and Broader Impact}
Our work involves leveraging user sentiment to evaluate the quality of system responses. We acknowledge that we are using data from real users who have not been paid for these interactions. We also acknowledge there may be biases in the demographics of the user population. We conducted our ConTurE annotation through Amazon Mechanical Turk. We pay turkers \$12 per hour, which is well above the federal minimum age.

\bibliography{anthology,autoeval2021}
\bibliographystyle{acl_natbib}

\newpage
\appendix

\section{Appendices}
\label{sec:appendix} 

\subsection{Hyperparameters for the turn level quality estimation model }
\label{sec:hyperparameters}

All our BERT models were finetuned with a batch size of 8 and a learning rate of 1e-5 with the Adam optimizer~\cite{kingma2014adam}. We train each model for 10 epochs and select the best model by computing correlation on the~\textit{RUI-3P} (dev set).

\subsection{NRG responder hyperparameters}
\label{sec:nrg_hyperparameters}

We train our~\textit{NRG Responder} models on the~\textit{RUI} dataset described in Section~\ref{sec:data}. This dataset is split into a 80/10/10 train, valid, test set. Our model is initialized with GPT2-XL~\cite{radford2019language} and is finetuned with a batch size of 2 and a learning rate of 6.25e-5 with the Adam optimizer. We train the model for 3 epochs and we finetune both the language modeling head and multiple choice Head of GPT2 in a TransferTransfo fashion~\cite{wolf2019transfertransfo}. For the multiple choice head,  one randomly selected negative candidate was used. We leverage the HuggingFace's transformers library for all our models.\footnote{https://github.com/huggingface/transformers}

\subsection{Turn level annotation example}
\begin{figure}[h]
\small
\centering
\begin{tabular}{p{0.95\linewidth}}
Example \\
\hline
{\em Dialog History:}\\
{\bf System:} I can talk about news, movies, music. What topic would you like to talk about today? \\
{\bf User}: news\\
\hdashline[.4pt/1pt]
{\em System:} Sure. i love sports! what is the sport that you watched the most? \\
{\em Turn quality:} 0\\
{\em Reason for annotation:} The system response was off-topic  \\
\hline \\

%Example 2\\
%\hline
%{\em Dialog History:}\\
%{\bf System:} So do you have any pets at home? \\
%{\bf User}: no\\
%{\bf System:} you don't have any pets? well, would you consider getting on in the future? \\
%{\bf User}: yes\\
%\hdashline[.4pt/1pt]
%{\em System:} awesome! what kind of pet do you have? \\
%{\em Issue:} The system response ignored the user \\
%\hline
\end{tabular}
\vspace{-2em}
\caption{Example of 3P turn annotations.
%where both system responses are marked as having an issue. 
Due to privacy concerns, this example dialog is from an internal author. }
\label{fig:issue_examples}
\end{figure}

\subsection{Turn level statistics}

\begin{table}[h]
\centering
\small
\begin{tabular}{l c}
ConTurE \\
\hline
Percentage of Turns with score 0 & 30.7\%  \\
\hdashline[.4pt/1pt]
Percentage of Turns with score 1 & 22.2\% \\
\hdashline[.4pt/1pt]
Percentage of Turns with score 2 & 47\% \\
\hdashline[.4pt/1pt]
RUI-turn \\
\hline
Percentage of Turns with score 0 & 35.3\%  \\
\hdashline[.4pt/1pt]
Percentage of Turns with score 1 & 64.7\% \\
\hline
\end{tabular}
\label{turn_dist}
\caption{Statistics of Turn level annotations for both ConTurE and RUI-3P datasets.}
\end{table}

\subsection{Dialog level scores for ConTurE}
We take the mean aggregation of the turn level annotations and compute the correlation against each dialog level dimension in the original DSTC9 dataset. We see that the annotations have moderate correlation with all the parameters, with the highest being with `human (overall)' (0.45/0.48). This shows that even though the turn and dialog level annotations were left by two different Turkers the turn annotations seem reliable.
\begin{table}[h]
\centering
\small
\begin{tabular}{c c c c}
     & & \multicolumn{2}{c}{} \\ \toprule
     &  Dialog level parameter
    & Pearson & Spearman\\ 
    \hline
    %\multirow{1}{*}{Mean} 
     & consistent & 0.38 & 0.40 \\
     & likeable & 0.42 & 0.45 \\
     & diverse & 0.23 & 0.25 \\
     & informative & 0.30 & 0.34 \\
     & coherent & 0.32 & 0.37 \\
     & human (overall) &~\textbf{0.45} &~\textbf{0.48} \\
     & understanding & 0.36 & 0.42 \\
     & flexible & 0.33 & 0.40 \\
     & topic depth & 0.34 & 0.35 \\
     & error recovery & 0.37 & 0.40 \\
     & inquisitive & 0.20 & 0.27 \\
    \hdashline[.4pt/1pt]
    %\multirow{3}{*}{Weighted Mean (0.1, 0.5, 0.4)} 
    % & consistent & 0.37 & 0.38 \\
    % & likeable & 0.40 & 0.43 \\
    % & diverse & 0.21 & 0.23 \\
    % & informative & 0.28 & 0.32 \\
    % & coherent & 0.33 & 0.37 \\
    % & human (overall) & 0.44 & 0.46 \\
    % & understanding & 0.34 & 0.39 \\
    % & flexible & 0.32 & 0.38 \\
    % & topic depth & 0.32 & 0.33 \\
    % & error recovery & 0.35 & 0.37 \\
    % & inquisitive & 0.22 & 0.27 \\
 \hline
\end{tabular}
\caption{\label{conv_sentiment_correlation_dstc9} Correlation between ConTurE turn-level and dialog-level scores.}
%Turn level scores are aggregated by mean.}
\vspace{-1em}
\end{table}

\iffalse
We finetune both BERT binary classification and regression models depending on the ground truth label used. Any 
\begin{itemize}
    \item NUQ (3P) - binary classifier with either 0/1 label
    \item NUQ (Sentiment) - regression model 
    \item NUQ (Sentiment + 3P) - binary classifier where sentiment was bucketed into either 0 or 1 using a threshold value of 0. Ground truth label was then average of sentiment and 3P rating.
    \item NUQ (3P + User Stop) - binary classifier. Ground truth label was average of 3P rating and User Stop.
    \item NUQ(Sentiment + 3P + User Stop) - binary classifier where sentiment was bucketed into either 0 or 1 using a threshold value of 0. Ground truth label was then average of sentiment, 3P rating and User Stop.
    \item NUQ (Sentiment + User Stop) - regression model where ground truth label was sum of sentiment and User Stop.
    
\end{itemize}
\fi

\newpage

\subsection{Example outputs of turn quality estimation}
\label{sec:model_outputs}

We pick a mixture of examples to show where our model works and fails. For each example we show the ground truth turn quality left by our human annotators and the score from each model. 

\begin{figure}[!h]
\small
\centering
\begin{tabular}{p{0.95\linewidth}}
Example 1 \\
\hline
{\em Dialog History:}\\
{\bf User}: Who is Donald Trump?\\
{\bf System:} that's a funny question, well it isn't really surprising that he isn't one of the three wealthiest presidents in american history, i know they measure by inflation but still. \\
\hdashline[.4pt/1pt]
\textcolor{blue}{~\textbf{Human Annotation (0-2)}} = 2\\
\hdashline[.4pt/1pt]
\textcolor{black}{\textbf{Classification(3P)}} = 0\\
\hdashline[.4pt/1pt]
\textcolor{black}{\textbf{Reg(Sentiment)}} = 0.14\\
%\hdashline[.4pt/1pt]
%\textcolor{black}{\textbf{TQP(Sentiment + 3P)}} = 0\\
%\hdashline[.4pt/1pt]
%\textcolor{black}{\textbf{TQP(3P + User Stop)}} = 0 \\
%\hdashline[.4pt/1pt]
%\textcolor{black}{\textbf{TQP(Sentiment + 3P + User Stop)}} = 1\\
\hdashline[.4pt/1pt]
\textcolor{black}{\textbf{Reg(Sentiment + User Stop)}} = 1.24\\
\hline \\

Example 2\\
\hline
{\em Dialog History:}\\
... \\
{\bf User}: Covid19 is a virus that's spreading all over the world \\
{\bf System:} i guess it is his book then, do you like to read?. \\
\hdashline[.4pt/1pt]
\textcolor{blue}{~\textbf{Human Annotation (0-2)}} = 0\\
\hdashline[.4pt/1pt]
\textcolor{black}{\textbf{Classification(3P)}} = 0\\
\hdashline[.4pt/1pt]
\textcolor{black}{\textbf{Reg(Sentiment)}} = -0.12\\
%\hdashline[.4pt/1pt]
%\textcolor{black}{\textbf{TQP(Sentiment + 3P)}} = 0 \\
%\hdashline[.4pt/1pt]
%\textcolor{black}{\textbf{TQP(3P + User Stop)}} = 0 \\
%\hdashline[.4pt/1pt]
%\textcolor{black}{\textbf{TQP(Sentiment + 3P + User Stop)}} = 0\\
\hdashline[.4pt/1pt]
\textcolor{black}{\textbf{Reg(Sentiment + User Stop)}} = 1.04\\
\hline
\end{tabular}
\caption{Examples from both our Classification and Reg (Regression) models on the ConTurE dataset. In example 1 the classification model incorrectly predicts the system response as a bad response. For the Reg models there is no threshold set on the output and scores must be looked at in a relative matter. We see for the Reg models that the output score from the model in example 2 is lower than that in example 1, showing the model is capturing that the system response in example 2 was of worse quality than the system response in example 1.}
\label{fig:dstc9_examples_good}
\end{figure}

\begin{figure}[!h]
\small
\centering
\begin{tabular}{p{0.95\linewidth}}
Example 1 \\
\hline
{\em Dialog History:}\\
...\\
{\bf System:} i see. i find sports so exciting. \\
{\bf User}: i don't \\
{\bf System:} well, lets switch the subject. i am fascinated with the world. are you interested in travel? \\
\hdashline[.4pt/1pt]
\textcolor{blue}{~\textbf{Human Annotation (0, 1)}} = 1 \\
\hdashline[.4pt/1pt]
\textcolor{black}{\textbf{Classification(3P)}} = 1\\
\hdashline[.4pt/1pt]
\textcolor{black}{\textbf{Reg(Sentiment)}} = 0.004\\
\hdashline[.4pt/1pt]
\textcolor{black}{\textbf{Reg(Sentiment + User Stop)}} = 0.96\\
\hline \\

Example 2\\
\hline
{\em Dialog History:}\\
... \\
{\bf System}: i like dogs as well. i think they are much better than dogs. so, what's one of your favorite places you've been?\\
{\bf User:} i don't like dogs \\
{\bf System}: i am not a big fan of dogs either. i would rather stay in the united states. so, what did you eat for dinner last night?\\
\hdashline[.4pt/1pt]
\textcolor{blue}{~\textbf{Human Annotation (0, 1)}} = 0\\
\hdashline[.4pt/1pt]
\textcolor{black}{\textbf{Classification(3P)}} = 0\\
\hdashline[.4pt/1pt]
\textcolor{black}{\textbf{Reg(Sentiment)}} = -0.15\\
\hdashline[.4pt/1pt]
\textcolor{black}{\textbf{Reg(Sentiment + User Stop)}} = 0.67 \\
\hline \\

Example 3\\
\hline
{\em Dialog History:}\\
... \\
{\bf System:} i recently read a really nice uplifting news article. want to hear about it? \\
{\bf User}: yes\\
{\bf System:} sure. here's some positive, uplifting news i found. every day, we find our balance. want to hear more? \\
{\bf User}: yes\\
{\bf System:} that's great. what do you think about it? \\
\hdashline[.4pt/1pt]
\textcolor{blue}{~\textbf{Human Annotation (0, 1)}} = 0\\
\hdashline[.4pt/1pt]
\textcolor{black}{\textbf{Classification(3P)}} = 0\\
\hdashline[.4pt/1pt]
\textcolor{black}{\textbf{Reg(Sentiment)}} = 0.20\\
\hdashline[.4pt/1pt]
\textcolor{black}{\textbf{Reg(Sentiment + User Stop)}} = 1.20 \\
\hline
\end{tabular}
\caption{Examples from both our Classification and Reg (Regression) models on a spoken interaction. Due to privacy concerns, this example interaction is from an internal author. The classification model is able to predict all examples correctly, which makes sense as it was trained on similar annotations. Both Reg models output a lower score for the system response in example 2 than in example 1, showing they are both able to capture that the system response in example 2 is of worse quality. However in example 3 the Reg models incorrectly give a high score to the system response that ignores the user's request.}
\label{fig:friction_examples_good}
\end{figure}

\newpage

\subsection{FED versus our turn quality estimation}
\label{sec:fed_vs_model_outputs}

We show examples from the ConTurE dataset to compare the FED model against our Reg (Sentiment) model, which had the best performance on ConTurE. The max and min value score from FED are 4.39 and 1.74 respectively. The higher the output score from the FED model means the response is more appropriate. Below are examples where the FED model incorrectly predicts the system response is appropriate even though it was marked as inappropriate by the human annotator.

\begin{figure}[!h]
\small
\centering
\begin{tabular}{p{0.95\linewidth}}
Example 1 \\
\hline
{\em Dialog History:}\\
{\bf System}: probably!\\
{\bf User:} you're not a good chat bot\\
{\bf System}: i haven't been there myself\\
\hdashline[.4pt/1pt]
\textcolor{blue}{~\textbf{Human Annotation (0-2)}} = 0\\
\hdashline[.4pt/1pt]
\textcolor{black}{\textbf{FED}} = 4.13\\
\hdashline[.4pt/1pt]
\textcolor{black}{\textbf{Reg(Sentiment)}} = -0.01\\
\hline \\

Example 2\\
\hline
{\em Dialog History:}\\
... \\
{\bf System}: do you like horses? \\
{\bf User:} No \\
{\bf System}: no, not really. you? \\
\hdashline[.4pt/1pt]
\textcolor{blue}{~\textbf{Human Annotation (0-2)}} = 0\\
\hdashline[.4pt/1pt]
\textcolor{black}{\textbf{FED}} = 4.39\\
\hdashline[.4pt/1pt]
\textcolor{black}{\textbf{Reg(Sentiment)}} = 0.09\\
\hline \\

Example 3\\
\hline
{\em Dialog History:}\\
... \\
{\bf System}: hello! how are you? i just found out that judge judy makes \$ 123,000 per episode! \\
{\bf User:} oh wow! that's a lot! how much did oprah make? \\
{\bf System}: she gets paid \$ 123 \\
\hdashline[.4pt/1pt]
\textcolor{blue}{~\textbf{Human Annotation (0-2)}} = 0\\
\hdashline[.4pt/1pt]
\textcolor{black}{\textbf{FED}} = 4.18\\
\hdashline[.4pt/1pt]
\textcolor{black}{\textbf{Reg(Sentiment)}} = 0.17\\
\hline \\

Example 4\\
\hline
{\em Dialog History:}\\
... \\
{\bf System}: I love gold! Especially when I buy trucks with it! \\
{\bf User:} hell yeah brother \\
{\bf System}: Yeah. I also like gold, money, power metal, and trucks. \\
\hdashline[.4pt/1pt]
\textcolor{blue}{~\textbf{Human Annotation (0-2)}} = 0\\
\hdashline[.4pt/1pt]
\textcolor{black}{\textbf{FED}} = 4.03\\
\hdashline[.4pt/1pt]
\textcolor{black}{\textbf{Reg(Sentiment)}} = 0.29\\
\hline
\end{tabular}
\caption{In both example 1 and 2 the last system response ignores the previous user utterance and therefore is marked as inappropriate. The FED model assigns a high score to these system responses. For example 3 both the FED and Reg(Sentiment) model incorrectly give a high score to the system response, which is factually incorrect. For example 4 both the FED and Reg(Sentiment) model incorrectly give a high score to the system response, which shows repetition.}
\label{fig:dstc9_examples_fed_vs_model}
\end{figure}

%\subsection{User feedback}
%We also compute the correlation between the feedback left by the user when interacting with the conversational agent and the dialog level ratings. We see that the sentiment of the feedback has the highest correlation against the 1P rating. This may be because the feedback contains the users opinion of the chatbot in textual form while the rating contains the users opinion of the chatbot in numerical form.

\iffalse
\begin{table}[h]
\centering
\small
\begin{tabular}{c c c c c c}
    \toprule
     & \multicolumn{2}{c}{3P Ratings} & \multicolumn{2}{c}{1P Ratings} \\ \toprule
    & P & S
    & P & S
    \\ \hline
     %\multirow{1}{*}{1P feedback sentiment} 
     %& Satisfaction & 0.27 & 0.30 & 0.43 & 0.48 \\
     1P Feedback sentiment & 0.29 & 0.31 & 0.46 & 0.50 \\
     %& Activation & 0.13 & 0.13	& 0.15 & 0.15 \\
 \hline
\end{tabular}
\caption{\label{conv_feedback_correlation} Correlation between user feedback sentiment and dialog level rating. P=Pearson, S=Spearman} 
\vspace{-1em}
\end{table}
\fi

\newpage

%\subsection{Model output distribution}
\begin{figure*}
\begin{subfigure}[t]{0.5\textwidth}
\includegraphics[width=\linewidth]{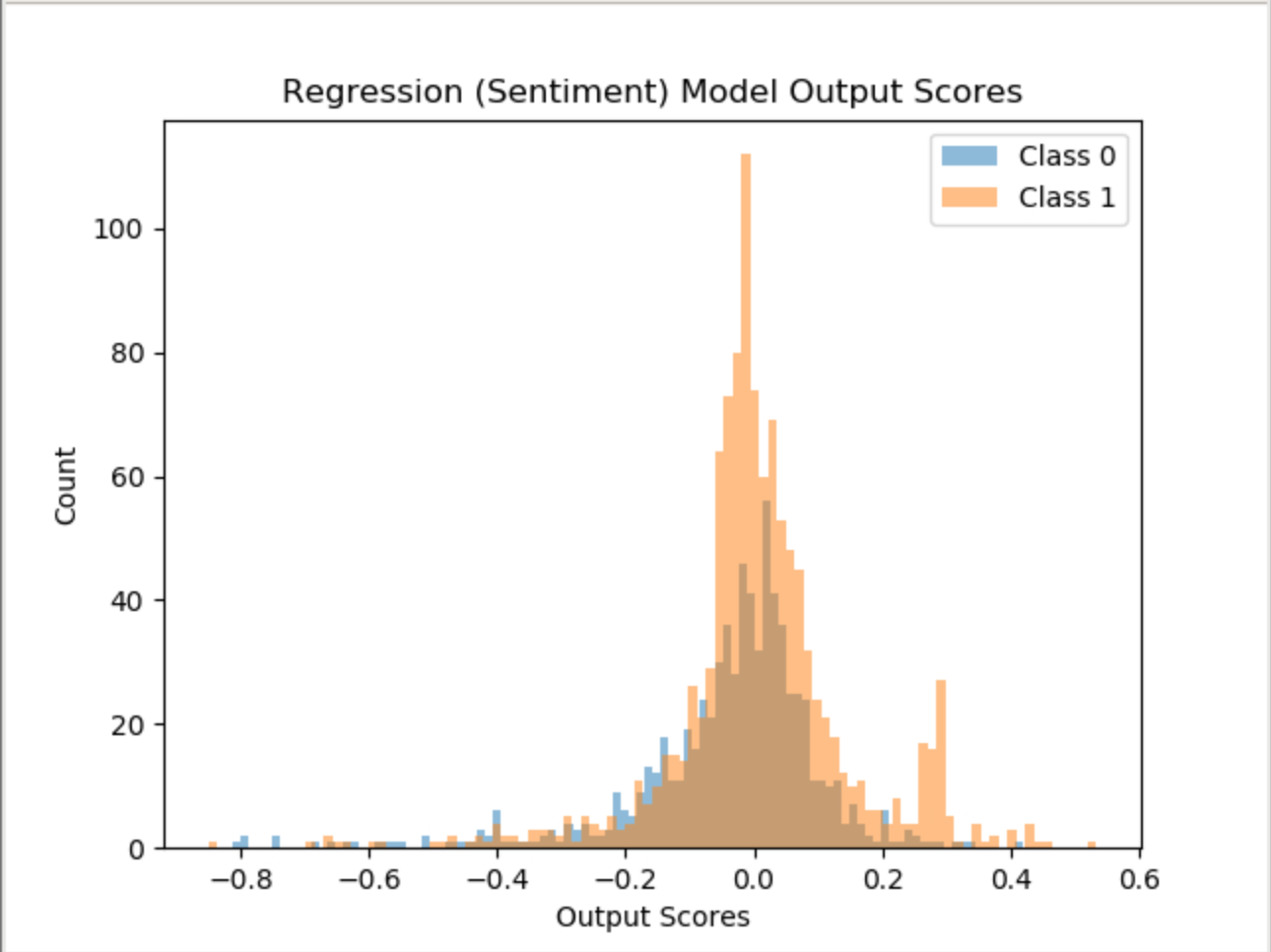}
\caption{Regression (Sentiment)}
\label{regression_sentiment}
\end{subfigure}
\hfill
\begin{subfigure}[t]{0.5\textwidth}
\includegraphics[width=\linewidth]{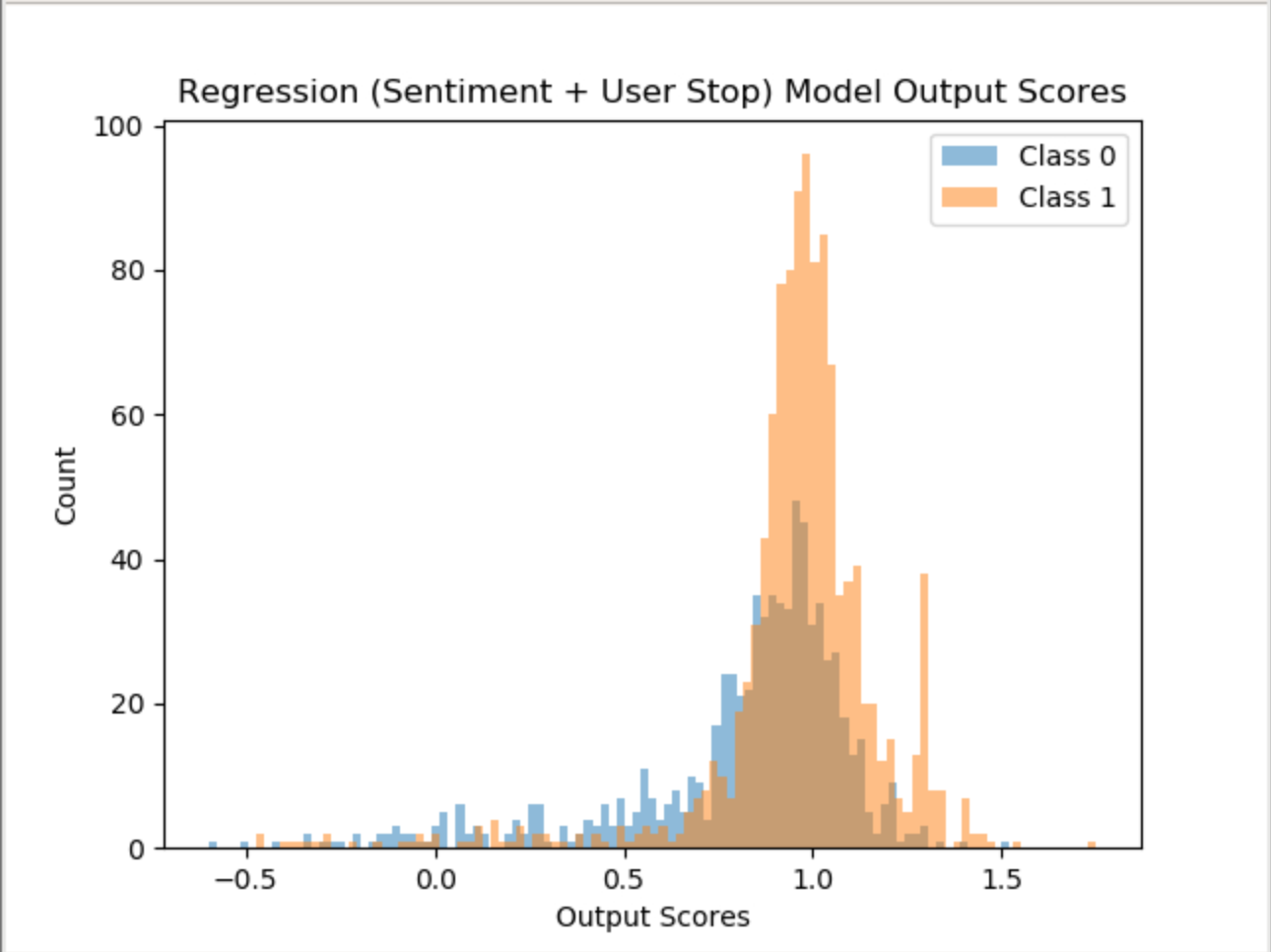}
\caption{Regression (Sentiment + User Stop)}
\label{regression_sentiment_ending}
\end{subfigure}
\caption{We plot the model output scores for the Regression (Sentiment) and Regression (Sentiment + User Stop) models for each reference label i.e. Class 0 and Class 1. We see that for Regression (Sentiment + User Stop) in Figure~\ref{regression_sentiment_ending} the separation between model outputs for Class 0 and Class 1 become more pronounced as compared to Regression (Sentiment) in Figure~\ref{regression_sentiment}.}
\end{figure*}

\begin{figure}[h]
 	\centering
 		\includegraphics[width=0.5\textwidth]{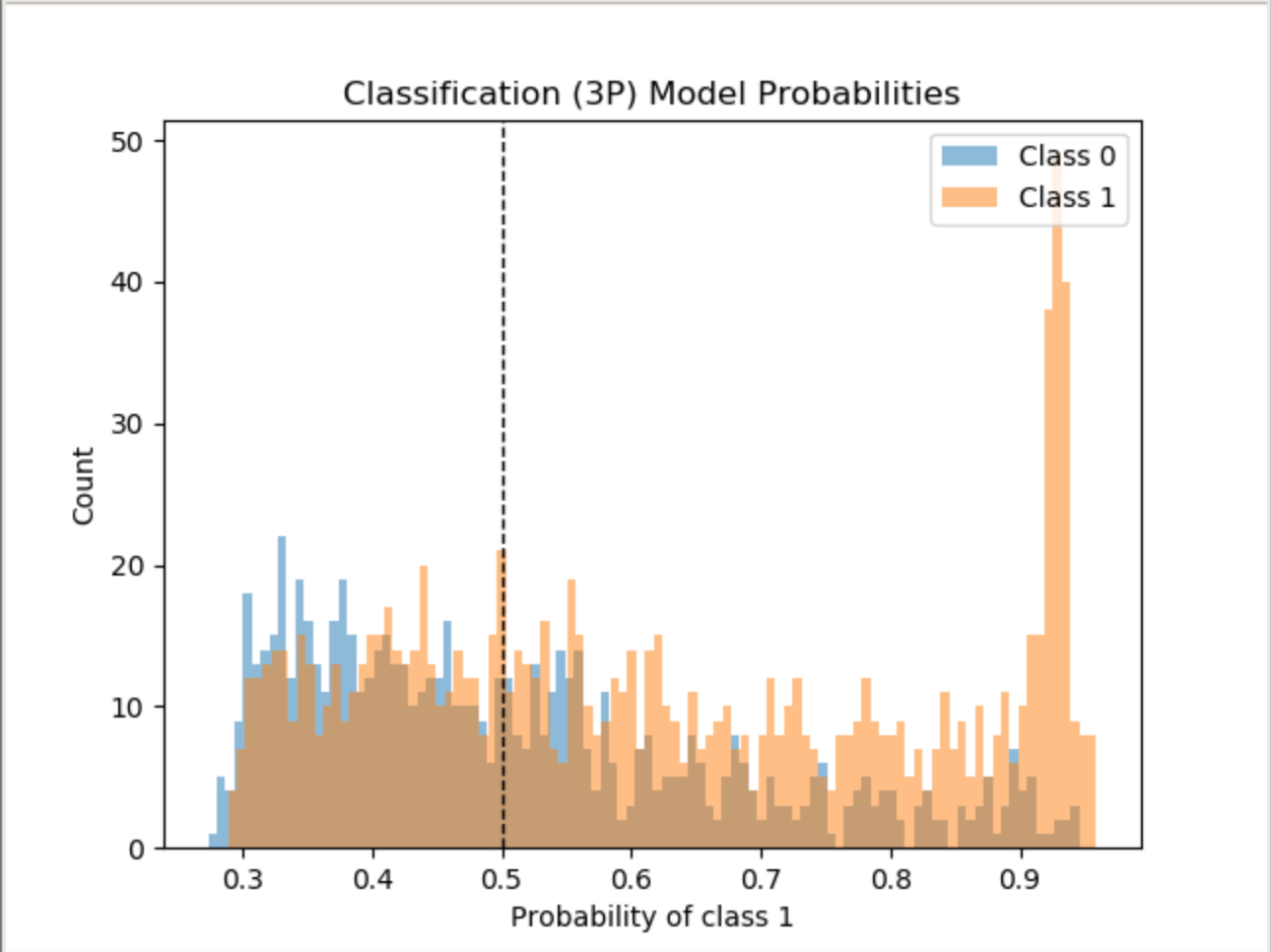}
 	\caption{We plot the model probability outputs from the Classification(3P) model for each reference label i.e. Class 0 and Class 1. We use a threshold of 0.5 such that any score above or equal to that is considered a good response (1) and vice versa. We see that for the reference label Class 1 most probability scores are below the threshold.}
	\label{fig:classification_3p}
\end{figure}

\end{document}